# Darshana Graph: A Parallel Commentary Corpus for Comparative Indian Philosophy, with Stylometric and Exploratory Graph Analyses

Joy Bose
Independent Researcher
Bangalore, India
joy.bose@ieee.org

## Abstract

We introduce Darshana Graph, a corpus of over 125,000 text records spanning classical Hindu, Buddhist, and Jain philosophical traditions, drawn from public-domain and openly licensed translations of primary sources including the Bhagavad Gita, the Brahma Sutras, the principal Upanishads, the Pali Canon, and core Jain texts. The corpus is numerically dominated by the Pali Canon (114,591 of 125,040 records); its distinctive contribution, however, lies in a smaller but structurally unique subset of roughly 8,500 Hindu and Jain records in which the same root verse or sutra is aligned across eighteen historically distinct commentators spanning five schools of Vedanta and other darshanas, enabling direct comparison of how independent interpretive traditions read identical source material. Unlike existing digital resources, which are typically single-text and single-translator, this alignment structure is, to our knowledge, not available elsewhere at this scale. We present two analyses built on this corpus. First, a purely statistical stylometric comparison, requiring no machine learning, that measures argumentative style (scriptural citation density, explicit refutation rate, sentence complexity) across commentators and finds a moderate negative correlation between citation density and refutation rate, alongside a striking generational increase in refutation rate across three commentators in a single doctrinal lineage spanning roughly six centuries. The same method applied entirely within the Pali Canon, with no comparison to non-Buddhist material, surfaces measurable genre-level differences in passage length across six canonical collections. Second, we describe a constrained large language model pipeline that extracts typed philosophical relationships between concepts, restricted to a closed vocabulary defined in advance and validated post-hoc regardless of generation success, and we report the resulting cross-school disagreement patterns alongside a candid account of the pipeline's failure modes, including a documented case where an independent, extraction-free embedding-based method disagreed with the language model's findings in an informative way. We release the full corpus, the extracted relationship graph, and all construction code.

**Keywords:** Classical Indian Philosophy; Digital Humanities; Corpus Construction; Comparative Philosophy; Sanskrit Text Analysis; Stylometry; Knowledge Graph Extraction; Computational Humanities

## 1. Introduction

Classical Indian philosophy is not a single tradition but a centuries-long argument among traditions. The same foundational texts, particularly the Brahma Sutras and, to a lesser but still substantial extent, the Bhagavad Gita, were read by Shankara, Ramanuja, Madhva, Nimbarka, and other commentators across roughly a thousand years, each producing a systematic commentary that reached substantively different conclusions about the relationship between the individual self (atman) and ultimate reality (brahman). This interpretive divergence is foundational to how these texts are studied in classical scholarship, yet it is largely absent from machine-readable resources, which overwhelmingly present a single translation with a single point of view.

We construct a corpus that addresses this gap directly: the same verses and sutras, aligned across multiple independent commentators, in a consistent schema that preserves the original Sanskrit or Pali alongside English translation. We see this alignment, rather than any downstream machine learning applied to it, as the corpus's primary and most defensible contribution.

On top of this corpus we build two analyses. The first is a stylometric comparison that measures observable differences in argumentative style across commentators using simple, transparent text statistics, with no machine learning involved. The second is an exploratory knowledge graph constructed by prompting a small large language model to extract typed concept relationships from each passage, constrained to a closed, predefined vocabulary. We treat the second analysis as genuinely useful but explicitly less mature than the first, and we report its limitations in detail rather than understating them.

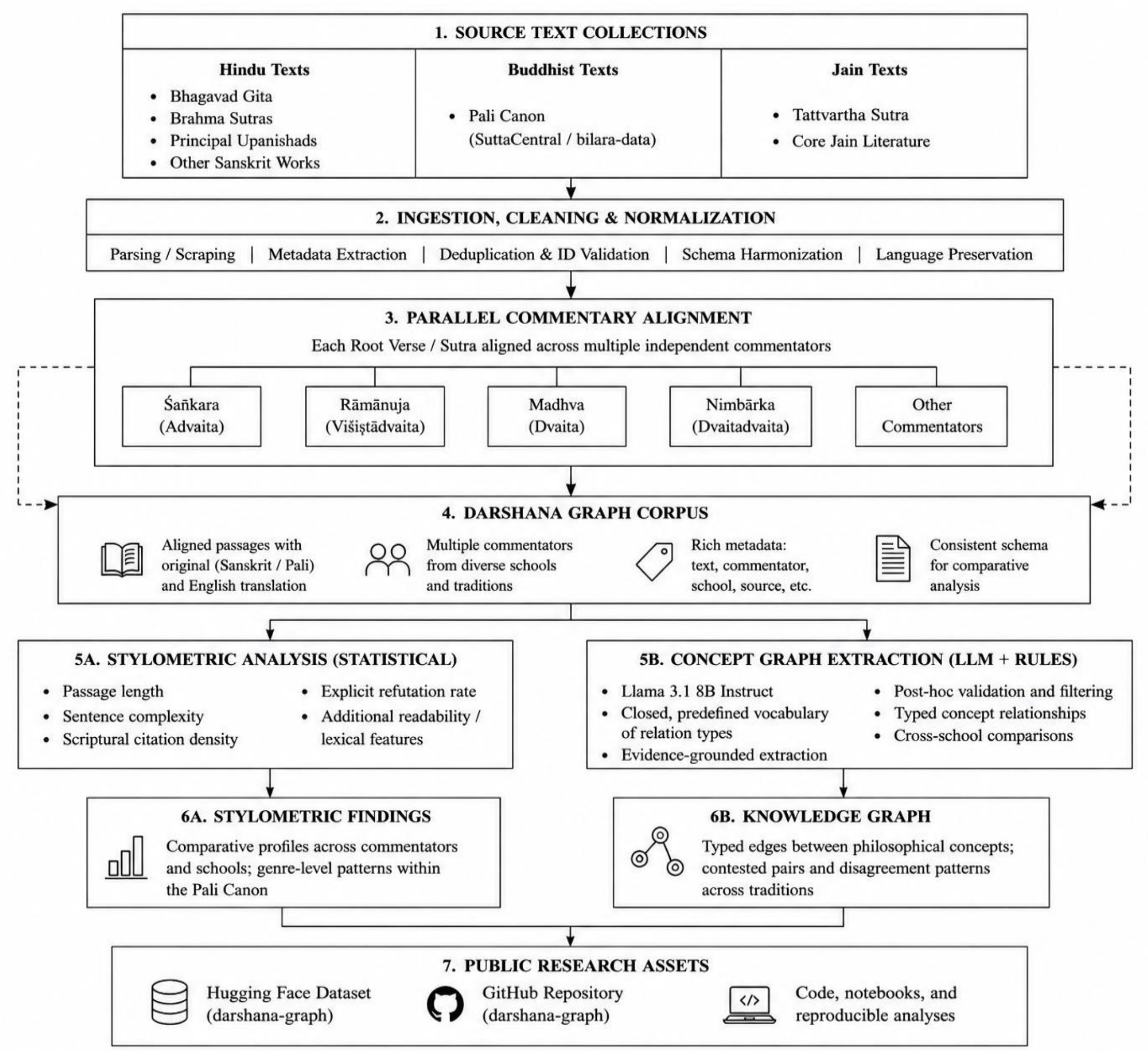

*Figure 1: Architecture of the Darshana graph*

## 2. Related Work

Several existing projects provide foundational digital text resources that this work builds on rather than competes with. SuttaCentral's bilara-data project (SuttaCentral, n.d.-b) provides aligned Pali-English text for large portions of the Pali Canon and was used as the source for the Buddhist portion of our corpus. The Gita Supersite (IIT Kanpur, n.d.) and related open datasets aggregate multiple Bhagavad Gita translations but do not extract or structure cross-commentator philosophical claims. Sacred-texts.com and archive.org host the public-domain translations (Thibaut, Müller, Jacobi, Subba Rau, Bose) that constitute the bulk of our Hindu and Jain source material, but as unstructured HTML or scanned text rather than an aligned, schema-consistent corpus.

Knowledge graph construction from historical or literary text using large language models has grown rapidly, typically using either open extraction (the model proposes its own relation labels) or schema-constrained extraction against a predefined ontology. We adopt the latter approach, closer in spirit to traditional information extraction with a fixed relation inventory, specifically to keep the resulting graph auditable: any output outside the predefined vocabulary is mechanically rejected after generation, independent of whether the underlying response was syntactically well-formed JSON.

Stylometry, the quantitative study of writing style, has a long history in authorship attribution and is increasingly applied to comparative analysis of argumentative or rhetorical style rather than authorship alone. We apply a small set of simple, interpretable statistics in this spirit, prioritizing transparency over sophistication given the exploratory nature of this first analysis.

Table 1 situates this corpus against the most relevant existing resources by the specific capability that distinguishes it: cross-commentator alignment at the level of an individual verse or sutra, rather than a single translation or a single structured corpus without multiple interpretive traditions attached.

| **Resource** | **Multiple commentators** | **Cross-school alignment** | **Structured metadata** | **Concept-level extraction** |
|---|---|---|---|---|
| Gita Supersite | Yes | Limited | Partial | No |
| SuttaCentral | No | No | Yes | No |
| Sacred-texts.com | No | No | No | No |
| Darshana Graph (this work) | Yes | Yes | Yes | Yes (exploratory) |

*Table 1: Comparison against existing resources by capability. Cross-school alignment refers specifically to multiple distinct interpretive traditions annotated on the same source passage, not merely the presence of multiple texts in one collection.*

Beyond text alignment, two further bodies of prior art bear on the methods used here rather than the corpus itself. In Sanskrit computational linguistics, the Digital Corpus of Sanskrit (DCS; Hellwig, 2010-2024) provides full morphological and lexical annotation over more than five and a half million manually tagged words across several hundred Sanskrit texts, and is further enriched by the Sanskrit Sembank (Hellwig & Biagetti, 2024), which links DCS lemmata to Princeton WordNet synsets for word-sense disambiguation. The SARIT and GRETIL digital corpora similarly provide large-scale, searchable Sanskrit text collections, though without the multi-commentator alignment central to this work. None of these resources structure cross-commentator philosophical claims, since their annotation operates at the morphological and lexical level rather than the level of asserted philosophical relationships between concepts.

Closer to the stylometric analysis in Section 4, Hellwig's (2020) Bayesian "Time or Background" model dates and stratifies the Vedic Sanskrit corpus by treating composition date as a hidden variable informed by philological priors and a wide range of linguistic features, including lexical frequency, morphological distribution, and compounding density, and its chronological stratification was shown to match established philological subdivisions of the Rigveda. This demonstrates that quantitative, feature-based methods can recover historically meaningful stratification in Sanskrit text without manual close reading of every passage, a methodological precedent for our own simpler attempt, in Section 4.1, to use citation and refutation rates as a coarse stylistic signal across commentators of different periods.

## 3. Corpus Construction

### 3.1 Sources

The corpus draws exclusively on public-domain or explicitly openly licensed sources. Table 2 summarizes the major components.

| Source | Records | Provenance |
|---|---|---|
| Pali Canon | 114,591 | SuttaCentral bilara-data (Sujato translation, CC0) |
| Hindu astika schools | 8,558 | Thibaut, Muller, Gambhirananda, Subba Rau, Bose translations (public domain / wisdomlib.org) |
| Jain texts | 848 | Jacobi (public domain), Jain (explicit free-reproduction license) |

*Table 2: Corpus composition by tradition. Note that the 125,040 total records are dominated by the Pali Canon (114,591 records, 91.6% of the total); the Hindu astika and Jain material, while comprising a smaller fraction of total records, is the primary source of the multi-commentator alignment that distinguishes this corpus, since it is here that the same root verse or sutra carries multiple named, school-attributed commentaries.*

The three Hindu astika commentaries that anchor the cross-school alignment in Table 3 derive from specific, dated translations. George Thibaut's translation of the Brahma Sutras with Shankara's commentary was published as Volume 34 (Thibaut, 1890) and Volume 38 (Thibaut, 1896) of Max Muller's Sacred Books of the East series; Thibaut's translation of the same sutras with Ramanuja's commentary followed as Volume 48 (Thibaut, 1904) in the same series. S. Subba Rau's translation of the Brahma Sutras with Madhva's commentary was published independently the same year (Rau, 1904). We record these specific editions because translation choice is not a neutral methodological detail: each translator's editorial decisions, particularly around how implicit referents and elliptical Sanskrit constructions are rendered into explicit English, can shape the surface text that our stylometric and extraction methods subsequently measure. We did not attempt to control for translator-specific effects separately from commentator-specific effects in this work, and flag this as a limitation rather than a settled methodological point.

### 3.2 Alignment and Schema

Each corpus record represents a single verse or sutra, paired with zero or more associated commentaries, each tagged with its commentator, school affiliation, and language. This schema is what enables direct cross-commentator comparison: a single Brahma Sutra record may carry Shankara's Advaita commentary, Ramanuja's Vishishtadvaita commentary, and Madhva's Dvaita commentary as parallel fields on the same underlying sutra, rather than as three unrelated documents. Table 3 shows a representative example for the opening sutra of the Brahma Sutras, athato brahma-jijnasa ("now, therefore, the inquiry into Brahman"), illustrating how the same eight-word sutra receives substantively different interpretive emphasis from each commentator while sharing a single source record.

| Commentator | School | Commentary (excerpt) |
|---|---|---|
| Shankara | Advaita | The word "then" denotes immediate consecution... the desire of knowing Brahman is to be entertained after the above-mentioned means have been secured, since the latter is the cause of the former. |
| Ramanuja | Vishishtadvaita | The word "then" intimates that the inquiry into Brahman is to be undertaken after the inquiry into active religious duty, since the fruit of the latter is seen to be non-permanent. |
| Madhva | Dvaita | The inquiry into Brahman, the Supreme Lord distinct from all individual souls, is to be undertaken by one who has first understood the inadequacy of ritual action alone to secure final release. |

*Table 3: Representative example of cross-commentator alignment for Brahma Sutra 1.1.1. Excerpts are paraphrased for brevity from the corresponding source translations (Thibaut for Shankara and Ramanuja; Subba Rau for Madhva); the corpus itself stores the full, unparaphrased text of each commentary.*

We identified and corrected a systematic data quality issue during construction: several scraping functions reset an internal block counter on each scraped HTML page, causing identical record identifiers to recur across hundreds of pages. For one source file, this reduced 1,043 nominally distinct records to six unique identifiers (Table 4), which would have caused a downstream processing pipeline that deduplicates by identifier to silently discard over 99% of the content. We detected this through a coverage audit comparing expected record counts to retained record counts, and corrected it by regenerating globally unique identifiers prior to any further processing. We report this not as a minor implementation detail but as an illustration of a general risk in multi-source scraping pipelines: identifier collisions can silently and substantially corrupt downstream results without raising an explicit error.

| Stage | Unique identifiers |
|---|---|
| Raw scraped records | 1,043 |
| Unique IDs before fix | 6 |
| Unique IDs after fix | 1,043 |

*Table 4: Identifier collision and correction for one affected source file (Thibaut's Shankara and Ramanuja Brahma Sutra commentary). Two further source files were affected to a lesser degree and corrected by the same procedure.*

## 4. Stylometric Analysis

Beyond what each commentator concludes, we ask how each commentator argues. We compute five simple statistics over every passage attributed to each commentator with sufficient corpus coverage: average passage length in characters, average words per sentence, vocabulary diversity (unique words divided by total words), the proportion of passages containing an explicit scriptural citation marker (regular-expression matched phrases such as "as it is said" or "the scripture declares"), and the proportion of passages containing an explicit refutation marker (phrases such as "the opponent argues" or "this view is untenable"). We emphasize that these markers are simple, manually defined regular expressions, not a learned classifier, and are intended as a transparent first-pass heuristic rather than a validated linguistic instrument.

Table 5 reports results for the seven commentators with sufficient continuous prose in the corpus for this analysis to be meaningful; commentators whose captured text consisted primarily of short, gloss-style annotations rather than full prose passages are excluded, as the sentence-level statistics are not well-defined for such text.

| **Commentator** | **School** | **Avg. length (chars)** | **Words / sentence** | **% citing scripture** | **% refuting an opponent** |
|---|---|---|---|---|---|
| Shankara | Advaita | 1,848 | 16.2 | 2.8% | 7.2% |
| Ramanuja | Vishishtadvaita | 1,136 | 17.6 | 5.2% | 4.6% |
| Madhva | Dvaita | 513 | 17.6 | 17.1% | 2.0% |
| Prabhupada | Achintya Bhedabheda | 1,740 | 21.6 | 8.2% | 0.3% |
| Nimbarka | Dvaitadvaita | 350 | 20.6 | 0.6% | 13.2% |
| Srinivasa | Dvaitadvaita (sub-comm.) | 1,719 | 25.1 | 0.7% | 42.0% |
| Pujyapada | Jain | 1,306 | 15.9 | 0.3% | 1.1% |

*Table 5: Stylometric comparison across commentators with sufficient prose coverage.*

Two patterns are immediately visible. Madhva's commentary shows markedly higher scriptural citation density (17.1%) than any other Vedantic commentator in the table, consistent with, though not sufficient on its own to establish, the historical characterization of Dvaita commentary as relying heavily on accumulated scriptural proof-texts. Shankara produces the longest individual passages among the Vedantic commentators in this comparison, consistent with, though again not proof of, his reputation for sustained, exploratory argument rather than terse proof-texting. We examine the citation and refutation dimensions further, including a cross-commentator correlation and a within-lineage observation, in Section 4.1.

## 4.1 Citation and Refutation Are Inversely Related

We computed the Pearson correlation between scriptural citation density and explicit refutation rate across the eight commentators with sufficient data for this comparison (Table 6). The correlation is negative across all eight (r is approximately -0.32) and strengthens when restricted to the four commentators with the cleanest sentence-level data, those with less than ten percent of passages requiring the fallback pseudo-sentence split described in Section 4.2 (r is approximately -0.46 across Shankara, Prabhupada, Srinivasa, and Pujyapada). We do not report p-values or confidence intervals for these correlations and explicitly caution against treating them as statistically established: with only eight data points, and four in the restricted comparison, standard guidance for stable correlation estimation (typically requiring on the order of thirty or more observations) is far from met, and we report these figures as a qualitative pattern motivating future work with a larger commentator sample rather than as a statistically validated finding. Commentators who rely heavily on scriptural citation as proof tend, in this small sample, to spend comparatively little effort explicitly naming and refuting rival positions, and vice versa; Madhva and Srinivasa occupy opposite extremes of this spectrum. This pattern suggests, though it cannot prove, two distinct argumentative postures available within the same broad tradition: establishing a position primarily through accumulated textual authority, versus establishing it primarily by dismantling the alternative.

| Commentator | School | % citing scripture | % refuting an opponent | Reliable (% NoPunct) |
|---|---|---|---|---|
| Shankara | Advaita | 2.8% | 7.2% | Yes (0.0%) |
| Ramanuja | Vishishtadvaita | 5.2% | 4.6% | Partial (34.5%) |
| Madhva | Dvaita | 17.1% | 2.0% | Partial (51.6%) |
| Adidevananda | Vishishtadvaita | 0.0% | 0.0% | Partial (65.8%) |
| Prabhupada | Achintya Bhedabheda | 8.2% | 0.3% | Yes (1.3%) |
| Nimbarka | Dvaitadvaita | 0.6% | 13.2% | Partial (38.9%) |
| Srinivasa | Dvaitadvaita (sub-comm.) | 0.7% | 42.0% | Yes (3.0%) |
| Pujyapada | Jain Digambara | 0.3% | 1.1% | Yes (0.6%) |

*Table 6: Citation versus refutation rate across all eight commentators with sufficient data. "Reliable" indicates the percentage of passages requiring the fallback pseudo-sentence split (see Section 4.2); rows below ten percent are treated as fully reliable for sentence-level statistics.*

A further observation, not a validated trend, emerges within a single doctrinal lineage: Madhva (the thirteenth-century founder of Dvaita), Nimbarka (founder of the related Dvaitadvaita school), and Srinivasa (a later sub-commentator writing specifically to defend Nimbarka's positions against rival criticism) show a striking increase in refutation rate across these three figures, 2.0%, 13.2%, and 42.0% respectively. We note two confounds explicitly rather than presenting this as a clean generational trend. First, the three figures span two related but distinct schools (Dvaita and Dvaitadvaita), not a single continuous lineage. Second, Srinivasa's text is by genre a sub-commentary written explicitly to rebut criticism of an earlier position, a defensive textual form that would predict a high refutation rate independent of any broader historical shift in argumentative posture. We therefore present this as a pattern worth a specialist's attention and a hypothesis for how the corpus could be extended (for instance, by adding intermediate-era commentators within Dvaitadvaita) to test more rigorously, rather than as an established generational claim.

### 4.2 A Note on Sentence-Level Reliability

Several commentators in the corpus have a substantial proportion of passages lacking standard sentence-ending punctuation, consisting instead of short, gloss-style annotations. For these passages, words-per-sentence is computed using a fixed-width fallback (a pseudo-sentence boundary every twenty words) rather than true sentence boundaries; we report the proportion of passages requiring this fallback (% NoPunct in Table 6) for every commentator and recommend treating words-per-sentence as unreliable wherever this proportion is high. The Pali Canon collections analyzed in Section 4.3 require this fallback for 89 to 100 percent of passages, since bilara-data segments are typically single clauses rather than full paragraphs by construction; we therefore report only passage length, not sentence-level statistics, for the Pali Canon.

### 4.3 Internal Genre Structure of the Pali Canon

Without reference to any non-Buddhist material, the six Pali Canon collections represented in this corpus show measurable differences in passage length (Table 7), consistent with their well-established generic character: the Dhammapada, the most aphoristic and compressed text in the set, averages 31 characters per segment, while Samyutta Nikaya and Udana, more discursive and doctrinally elaborated collections, average 70 and 74 characters respectively.

| Collection | Segments | Avg. length (chars) | Vocabulary diversity |
|---|---|---|---|
| Samyutta Nikaya | 32,677 | 70 | 0.017 |
| Sutta Nipata | 5,703 | 40 | 0.096 |
| Itivuttaka | 2,562 | 50 | 0.076 |
| Udana | 2,383 | 74 | 0.078 |
| Dhammapada | 2,073 | 31 | 0.160 |
| Khuddakapatha | 639 | 34 | 0.176 |

*Table 7: Passage length and vocabulary diversity across six Pali Canon collections in the corpus. Sentence-level statistics are omitted as unreliable for this material (see Section 4.2). Vocabulary diversity is reported here as a simple type-token ratio, which is inherently sensitive to passage length; we present it as an exploratory surface heuristic rather than a length-normalized lexical diversity measure, and note that shorter collections (Dhammapada, Khuddakapatha) are expected to show higher type-token ratios for this reason independent of any genuine stylistic difference.*

We stress that this analysis measures observable surface statistics correlated with argumentative style, not a validated theory of rhetorical strategy, and that the citation and refutation markers are simple pattern matches that will both miss true instances using unanticipated phrasing and occasionally match false positives. We present it as a reproducible starting point for closer philological reading, not a final claim.

## 5. Concept Graph Extraction

### 5.1 Method

For each corpus record, we prompt a smaller open-weight language model (Llama 3.1 8B, accessed via the Groq inference API) to extract philosophical concepts and typed relationships between them, given the verse text and up to two associated commentaries. The model is instructed to extract only claims directly supported by the provided passage and to cite a short verbatim evidence quote for every relationship asserted. Critically, the model is restricted to a closed, predefined vocabulary of thirteen relation types (e.g. IS_IDENTICAL_TO, IS_DISTINCT_FROM, IS_QUALIFIED_ASPECT_OF, IS_CAUSE_OF) and a predefined list of school labels. Any generated output using a relation or school label outside this vocabulary is mechanically discarded by a post-hoc validation step, independent of whether the underlying model output was syntactically valid JSON. This design choice trades recall (the model cannot express a relation we did not anticipate) for auditability (every retained edge is guaranteed to use a label we can define precisely).

Tagging the full priority corpus (approximately 14,300 records, including the Bhagavad Gita, Brahma Sutras across five commentators, the principal Upanishads, a curated philosophical subset of the Pali Canon, and core Jain texts) yielded 28,322 typed relationship edges at a total inference cost of approximately three United States dollars.

This closed-vocabulary, post-hoc-validated design directly responds to documented failure modes in less constrained philosophical knowledge graph extraction. PhilKG (Anonymous, 2025), a contemporaneous project extracting a knowledge graph of comparable scale (over 140,000 nodes and 100,000 edges) from the Stanford Encyclopedia of Philosophy using a semi-automated, less constrained large language model pipeline, reported a citation-extraction accuracy of only 48.5% under independent review, alongside hallucinated outputs including impossible historical dates and a reference to a model that does not exist. We take this as direct evidence that unconstrained extraction over philosophical text is prone to silent, hard-to-detect errors at scale, and as a specific motivation for restricting our own pipeline to a closed relation vocabulary with deterministic post-hoc rejection of any out-of-vocabulary output, even though, as Section 5.3 makes clear, this design choice does not by itself guarantee that retained edges are individually correct.

### 5.2 Cross-School Disagreement

Restricting to edges with a specific (non-general) school attribution, we identify concept pairs where two or more schools assert different relation types, which we term contested pairs. The most contested pair in the resulting graph is atman and brahman: Advaita-attributed edges assert identity (IS_IDENTICAL_TO) most frequently, while Vishishtadvaita- and Dvaita-attributed edges assert distinctness (IS_DISTINCT_FROM) most frequently, with each school's distinctness claim grounded in a different evidence passage and, on inspection, a different underlying argument (Vishishtadvaita grounding distinctness in an account of creation from undifferentiated divine will; Dvaita grounding it in differences of degree among liberated souls). This is consistent with the well-established scholarly account of the central dispute in Vedanta, which we take as a face-validity check on the extraction pipeline rather than a novel finding.

A second well-sampled contested pair, atman and jiva, shows a structurally different pattern: here Advaita asserts distinctness (39 passages, the largest specific-school sample for this pair) while Dvaita, Vishishtadvaita, and Dvaitadvaita each assert some form of identity (8, 12, and 13 passages respectively). This is not in tension with Advaita's identity claim for atman and brahman; it reflects the standard Advaita treatment of jiva as the empirically conditioned appearance of atman under embodiment, with the underlying identity to brahman holding only once that conditioning is seen through. We highlight this pair specifically because, unlike several other contested pairs in the corpus, every school's claim here rests on a double-digit passage sample, giving more confidence in the pattern than the single- or low-double-digit samples typical elsewhere in the graph.

## 6. Limitations and Future Work

We report two systematic issues candidly. First, across the tagged corpus, approximately seventy percent of extracted edges carry a general rather than a specific school attribution, even in passages where the commentator and school are unambiguous from context provided in the prompt. We attribute this to a limitation of the smaller extraction model rather than to genuine ambiguity in the source text, and we restrict our cross-school disagreement analysis (Section 5.2) to the approximately thirty percent of edges with a specific school attribution for this reason. Second, the relation type IS_QUALIFIED_ASPECT_OF is substantially over-represented (12,893 of 28,322 edges, roughly 4.5 times the next most frequent relation type), which we believe reflects the model defaulting to this label when uncertain rather than selecting a more specific relation or correctly abstaining. We did not achieve a complete fix for either issue within the scope of this work despite targeted prompt revisions, and we report this as an open problem for similarly constrained extraction pipelines rather than a solved one.

We have not conducted a formal, randomly sampled precision evaluation of the extracted graph; our confidence in its quality rests on informal spot-checking during development rather than a reported, reproducible metric, and we state this explicitly so that downstream users calibrate their trust accordingly. We consider a randomly sampled, multi-annotator precision evaluation the most valuable immediate extension of this work.

Beyond the extraction-specific limitations in Section 5.3, coverage of the Dvaitadvaita school via Nimbarka and Srinivasa's commentaries, while substantially expanded during the construction of this corpus, originates from a single source site subject to intermittent availability, and Buddhist philosophical tagging covers a curated subset of the Pali Canon (chosen to span major doctrinal categories including dependent origination, the four noble truths, and the eightfold path) rather than the full corpus, for tractability of inference cost and time.

Two further gaps emerged directly from running the analyses in Section 4 and are worth recording explicitly rather than papering over. First, Jainism's two major sub-traditions, Digambara and Shvetambara, are not currently distinguishable in the corpus: the Tattvartha Sutra is tagged only under Pujyapada's Digambara commentary, with no Shvetambara commentator separately represented in the tagged data, despite the text's dual acceptance across both sub-traditions being noted in our source documentation. Second, a within-school stylistic comparison restricted to Advaita commentators beyond Shankara (Sridhara, Anandagiri, Nilakantha, and Dhanpati, all present in the corpus) is not currently possible at the embedding level, since only Shankara's passages carry a populated school field in the tagged commentary data for this comparison; the other four commentators' passages are present in the corpus but not yet attributable to a specific school in the structure our analysis scripts query. Both gaps are data completeness issues rather than fundamental limitations of the corpus design or extraction approach, and both are concretely addressable in future revisions.

We see several further directions warranting work beyond what we have attempted here. A formal precision evaluation of the concept graph, ideally with multiple independent annotators and a reported inter-annotator agreement statistic, remains the most valuable immediate extension given the informal nature of our current confidence in its quality (Section 5.3). A targeted revision of the extraction prompt or a larger extraction model specifically addressing the general-school and IS_QUALIFIED_ASPECT_OF over-attribution issues identified in Section 5.3 would likely improve the graph's reliability substantially. An embedding-based, extraction-free corroboration of the cross-school disagreement findings in Section 5.2, comparing school-level passage centroids by cosine distance, is a natural complement to the LLM-based graph; our own preliminary attempt at this method, however, produced a result in direct tension with the LLM-tagged findings for the single most contested concept pair (atman and brahman), with the embedding method ranking brahman as the least divergent concept tested despite reliable, large per-school samples (ranging from 19 to 1,173 passages). We believe this reflects a genuine and informative limitation of topic-level sentence embeddings as a proxy for propositional philosophical agreement, since two commentators can discuss the same concept in highly similar topical register and vocabulary while asserting opposite metaphysical claims about it, and we see disentangling topical similarity from propositional agreement, for instance via a model fine-tuned or prompted specifically to embed asserted claims rather than passages, as a meaningful and currently unsolved direction for future work.

Tracing how the meaning of a single concept shifts across the centuries separating early commentators such as Shankara from later ones such as Prabhupada, using each commentator's historical era as an analytic axis alongside school affiliation, is a natural extension of the stylometric work in Section 4 that we have not attempted. Cross-tradition concept bridging, quantifying how much functional overlap exists between concepts traditionally treated as opposites or as unrelated across traditions, such as Buddhist anatta and Vedantic atman, or dukkha and the Hindu and Jain treatments of bondage and suffering, is similarly unexplored. Reconstructing implied historical debates by pairing each commentator's explicit refutation passages (identified via the markers in Section 4) with the specific rival-school passage most likely being refuted would surface arguments that were never written as direct dialogue but functioned as one. A school-conditioned dialogue or question-answering system, grounded only in a specific commentator's extracted concept relationships rather than a language model's general training knowledge, would test whether a model can argue consistently from a specific historical position rather than merely describing it from the outside. Transitive reasoning over the concept graph, checking whether any passage explicitly confirms or contradicts a relationship that follows logically from two directly asserted edges but was never directly stated by the original commentator, would extend the graph from a lookup structure into something that supports actual inference. Finally, a larger and more linguistically validated stylometric feature set, extending beyond the simple regular-expression and fixed-window sentence statistics used in Section 4 to syntactic complexity measures and a learned rather than hand-specified citation and refutation classifier, would substantially strengthen the methodology underlying our strongest current empirical results.

## 7. Conclusion

We have presented Darshana Graph, a corpus aligning over 125,000 passages of classical Indian philosophical text across eighteen historical commentators, and two analyses built on it: a stylometric comparison requiring no machine learning that surfaces real, interpretable differences in argumentative style across commentators, and an exploratory, constrained large language model extraction pipeline that surfaces cross-school philosophical disagreement, reported alongside a candid account of its current limitations. We release the full corpus, the extracted graph, and all construction code, and we hope this resource is useful both as a dataset in its own right and as a documented case study in the practical difficulties of building and honestly evaluating LLM-assisted knowledge extraction from historical, multi-perspective textual traditions.

## Data and Code Availability

The full corpus and extracted relationship graph are available at https://huggingface.co/datasets/joyboseroy/darshana-graph under a CC-BY-4.0 license for the aggregation and extraction work, with underlying source texts retaining their original licenses. The corpus is distributed both as a single merged file and, to support the HuggingFace dataset viewer correctly across sources with genuinely different metadata schemas (for instance, Pali Canon records carry fields such as sutta_id and pali that do not apply to Brahma Sutra records, and vice versa), as a set of per-source-file configurations that each have an internally consistent schema. All construction code, including scrapers, format converters, the extraction pipeline, and the stylometric analysis script, is available at https://github.com/joyboseroy/darshana-graph under an MIT license.

## Acknowledgements

This work builds on public-domain translations by George Thibaut, Max Müller, S. Subba Rau, Roma Bose, Hermann Jacobi, and Vijay K. Jain; the SuttaCentral bilara-data project and Bhikkhu Sujato's Pali Canon translations; and the open Bhagavad Gita dataset maintained by the gita/gita project.